\documentclass[conference]{IEEEtran}
\def\IEEEtitletopspace{22pt}
\IEEEoverridecommandlockouts

\usepackage{cite}
\usepackage{amsmath,amssymb,amsfonts}
\usepackage{algorithmic}
\usepackage{graphicx}
\usepackage{textcomp}
\usepackage{xcolor}
\usepackage{placeins}

\usepackage[top=50pt, left=46pt, right=46pt, bottom=54pt]{geometry}

\def\BibTeX{{\rm B\kern-.05em{\sc i\kern-.025em b}\kern-.08em
		T\kern-.1667em\lower.7ex\hbox{E}\kern-.125emX}}

\usepackage{booktabs}
\usepackage{multirow}

\graphicspath{{./figures/}{./images/}{./}}

\begin{document}

\title{\textbf{Contact Status Recognition and Slip Detection with a Bio-inspired Tactile Hand} \\
\thanks{This work was supported by the State Key Laboratory of Autonomous Intelligent Unmanned Systems (ZZKF2025-2-5) and the State Key Laboratory of Robotics and Systems (HIT) (SKLRS-2025-KF-11). (Corresponding author: Longhui Qin, lhqin@seu.edu.cn).}
}

\author{\IEEEauthorblockN{Chengxiao He}
	\IEEEauthorblockA{\textit{School of Mechanical Engineering, Southeast University}\\
		Nanjing 211189, China}
	\and
	\IEEEauthorblockN{Wenhui Yang}
	\IEEEauthorblockA{\textit{School of Mechanical Engineering, Southeast University}\\
		Nanjing 211189, China}
	\and
	\IEEEauthorblockN{Hongliang Zhao}
	\IEEEauthorblockA{\textit{School of Mechanical Engineering, Southeast University}\\
		Nanjing 211189, China}
	\and
	\IEEEauthorblockN{Jiacheng Lv}
	\IEEEauthorblockA{\textit{School of Electronic Science \& Engineering, Southeast University}\\
		Nanjing 211189, China}
	\and
	\IEEEauthorblockN{Yuzhe Shao}
	\IEEEauthorblockA{\textit{School of Mechanical Engineering, Southeast University}\\
		Nanjing 211189, China}
	\and
	\IEEEauthorblockN{Longhui Qin$^{*}$}
	\IEEEauthorblockA{\textit{School of Mechanical Engineering, Southeast University}\\
		Nanjing 211189, China}
}

	
\maketitle

\begin{abstract}
\bfseries\boldmath
Stable and reliable grasp is critical to robotic manipulations especially for fragile and glazed objects, where the grasp force requires precise control as too large force possibly damages the objects while small force leads to slip and fall-off. Although it is assumed the objects to manipulate is grasped firmly in advance, slip detection and timely prevention are necessary for a robot in unstructured and universal environments. In this work, we addressed this issue by utilizing multimodal tactile feedback from a five-fingered bio-inspired hand. Motivated by human hands, the tactile sensing elements were distributed and embedded into the soft skin of robotic hand, forming 24 tactile channels in total. Different from the threshold method that was widely employed in most existing works, we converted the slip detection problem to contact status recognition in combination with binning technique first and then detected the slip onset time according to the recognition results. After the 24-channel tactile signals passed through discrete wavelet transform, 17 features were extracted from different time and frequency bands. With the optimal 120 features employed for status recognition, the test accuracy reached 96.39\% across three different sliding speeds and six kinds of materials. When applied to four new unseen materials, a high accuracy of 91.95\% was still achieved, which further validated the generalization of our proposed method. Finally, the performance of slip detection is verified based on the trained model of contact status recognition. 
\end{abstract}


\section{Introduction}
Relying on tactile feedback from the fast-adapting (FA) and slow-adapting (SA) types of mechanoreceptors within human skin, we are able to perceive kinds of stimuli from external environments and manipulate various objects dexterously. As for robotic manipulations, tactile perception is also required in complex tasks to regulate grip forces, maintain stable contacts and react to incipient slip \cite{roberts2021soft, zhan2025recent, xin2025vision_based_tactile}. To ensure firm grasp, the contact force should be precisely controlled within a reasonable range, especially for delicate manipulations of fragile or glazed objects, in which slip detection becomes significant to avoid falling off of in-hand objects.

\begin{figure}[!t]
	\centering
	\includegraphics[width=0.9\columnwidth]{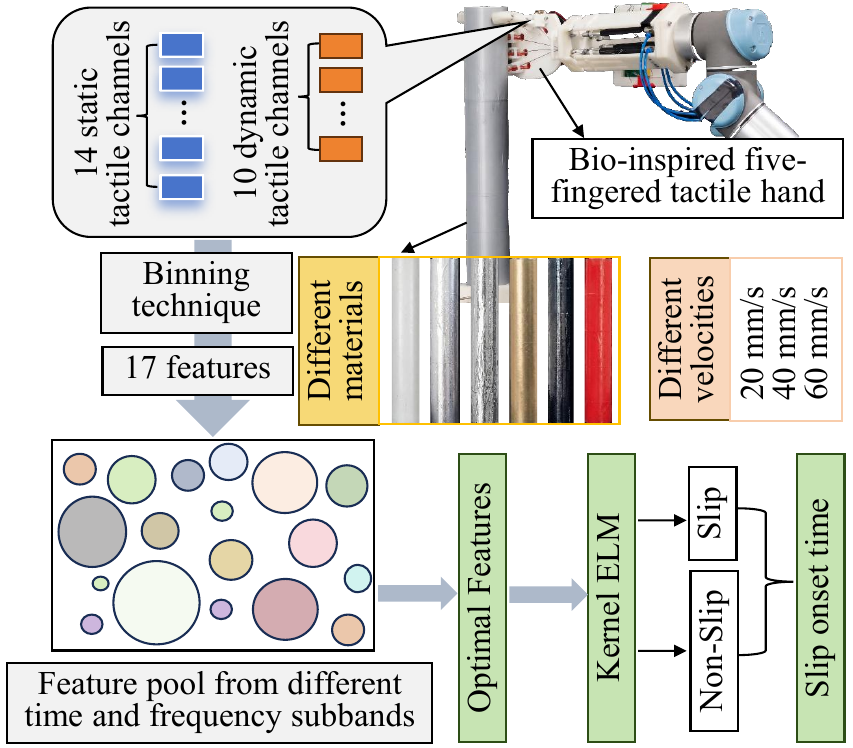}
	\caption{Illustration of the core idea of our work. The 24-channel tactile signals generated in the bio-inspired hand are processed via our proposed signal framework to maximally dig out the contained valid information. Slip onset time can be determined according to the contact status recognition result. Two key factors, i.e., materials and sliding velocities are also considered.}
	\label{fig:graphic_abstract}
\end{figure}

In the majority of existing works on slip detection, a two-finger gripper is employed \cite{wang2021gelsight, sui2022incipient} and the threshold comparison method is widely adopted \cite{romeo2020slip_survey, romeo2021automatic_slippage}, which defines an empirical threshold in advance and slippage is detected once the measured value exceeds this threshold. Although it is advantageous in simple implementation and fast calculation speed, it poses strict requirements for the sensitivity of perception signals and the level of signal-to-noise ratio. More importantly, it is hard to provide generalized performance, i.e., it most likely fails when the materials of grasped objects change. Another prevailing approach for slip detection in recent years is learning-based method, which learns the characteristics of slippage onset time via the training of a neural network (NN) \cite{agriomallos2018slippage, chen2025learning}. It outperforms the conventional comparison-via-threshold method due to more generalized applicability. Before feeding into a classification model, the features are usually extracted via hand-crafted features or off-the-shelf convolutional neural network (CNN) models. However, it stays challenging how to dig out the optimal features for the targeted problem and transfer a trained model across different materials.

In this paper, we tried to address this issue by developing a bio-inspired tactile hand with 14 piezoresistive sensing elements (SEs) and 10 piezoelectric SEs embedded into the soft skins (Fig. \ref{fig:graphic_abstract}). The two types of tactile channels correspond to the SA and FA types of mechanoreceptors in human hand, as they are respectively responsible for the measurement of static and dynamic forces. A signal processing framework was proposed to recognize the non-slip and slip contact status and then detect the slip onset time. It mainly consists of signal preprocessing, decomposition and transformation, rapid extraction of the optimal features and binary classification. To improve the adaptability in general cases, three different sliding speeds and six materials were employed to train the recognition model. An accuracy as high as 96.39\% was achieved. It was further verified in four unseen materials and the accuracy reached 91.95\%. The performance of our proposed design and method was finally demonstrated with the potential it had shown in determining the exact slip onset time. 

The contribution of our work can be summarized as: 1) We developed a bio-inspired five-fingered tactile hand comprising 14 static and 10 dynamic tactile channels in the soft skin. 2) A signal processing framework tailored for slip detection was proposed to maximally leverage the multi-channel tactile information. 3) The effectiveness of our method was validated through robotic experiment and its generalization was successfully demonstrated on unseen materials.

\section{Related Works}

\subsection{Tactile Hand/Finger Designs for Slip Perception}
Slip perception in robotic manipulation requires capturing both quasi-static contact and high-frequency micro-vibration signatures. According to the working principle, there have been a variety of tactile sensors for robotics, including resistive/capacitive, piezoelectric/triboelectric, magnetic, and vision-based sensors \cite{zhan2025recent,roberts2021soft}. A widely adopted end-effector is two-finger parallel grippers where the tactile sensors were usually attached to the surface of a fingertip for perception \cite{wang2021gelsight,sui2022incipient, qin2025recent}. 
Vision-based tactile sensors (VBTS) offer dense deformation or marker-flow fields, providing rich slip-related information, particularly beneficial for detecting incipient micro-slips \cite{wang2021gelsight, lambeta2020digit}. It could be integrated into different designs for slip detection \cite{xin2025vision_based_tactile, sui2022incipient}. 

By contrast, multi-fingered robotic hands exhibited more advantages due to increased flexibility, enlarged contact region and enhanced perception sensitivity \cite{egli2024sensorized}. The most usual manner to empower a robotic hand with tactile perception is to attach electronic skin to its surface. The development of energy-efficient and self-powered electronic skins (e-skins) addressed challenges associated with power consumption and wiring complexity in large-area tactile sensing systems \cite{egli2024sensorized, xu2024selfpowered_eskin}. Although advancements in soft and stretchable sensor materials have enabled higher conformity to curved surfaces \cite{li2020skin_inspired_quadruple, min2025stretchable}, there are still other issues to consider in this manner, such as the reliability, cost and durability. Multimodal tactile sensors that combined strain gauge (SG) with PVDF (Polyvinylidene fluoride) could also be applied to realize slip recognition \cite{gao2024aim_bionic_tactile}. In capturing vibration-based slip cues, piezoelectric materials, such as PVDF, were particularly advantageous due to their high bandwidth, sensitivity, and straightforward readout mechanisms \cite{park2025manufacturing, qin2023perception, zhao2025multi_perspective_feature}. However, the majority of existing designs were only equipped with the tactile perception capability on the fingertip, which was limited in practical applications. In our design, multiple tactile SEs were embedded into the soft skin layer of different segments of a finger and all five fingers possessed the perception capability.

\subsection{Slip Detection Algorithms}
The conventional algorithm for slip detection is threshold comparison, in which the signals are analyzed in advance and the signal magnitude corresponding to slip onset time is determined as the threshold. Once the filtered vibration amplitudes or force derivatives reach or exceed the preset threshold, the time instant is thought as the slip onset \cite{romeo2020slip_survey,romeo2021automatic_slippage}. In \cite{liu2024ultrasensitive}, slip was detected once the maximal discrete wavelet transform (DWT) detail coefficient exceeded the slipping threshold. It is straightforward and convenient to implement in practical scenarios. However, it suffers from extremely high requirement for the sensitivity of tactile sensors and close relationship to the physical properties of object surfaces.

Recent advancements have explored richer temporal and spatial modeling approaches, including graph-based methods that fused tactile readings distributed across multiple finger locations \cite{funabashi2022gcn_tactile} and cross-modal techniques that integrated complementary sensing streams to improve robustness and generalizability \cite{chen2025crossmodal_slip}. Additionally, some approaches leveraged deformation-field analysis while some others realized physics-informed modeling, combining contact mechanics and slip-consistency constraints explicitly\cite{jawale2024learned, hu2024learning}. Although automatic feature extraction by CNN eases signal processing and reduces human intervention, hand-crafted features that customized for specific problem still showed advantages in the prediction accuracy \cite{zhao2025multi_perspective_feature}. In this paper, we converted the slip detection problem into contact status recognition first and then proposed the binned 24-channel tactile signals with selected optimal features and extreme learning machine (ELM), which was different from existing methods and had been demonstrated with robotic explorations.

\section{Tactile Hand and Experimental Setup}
\subsection{Bio-inspired Tactile Hand}
\begin{figure}[htbp]
	\centering
	\includegraphics[width=0.9\columnwidth]{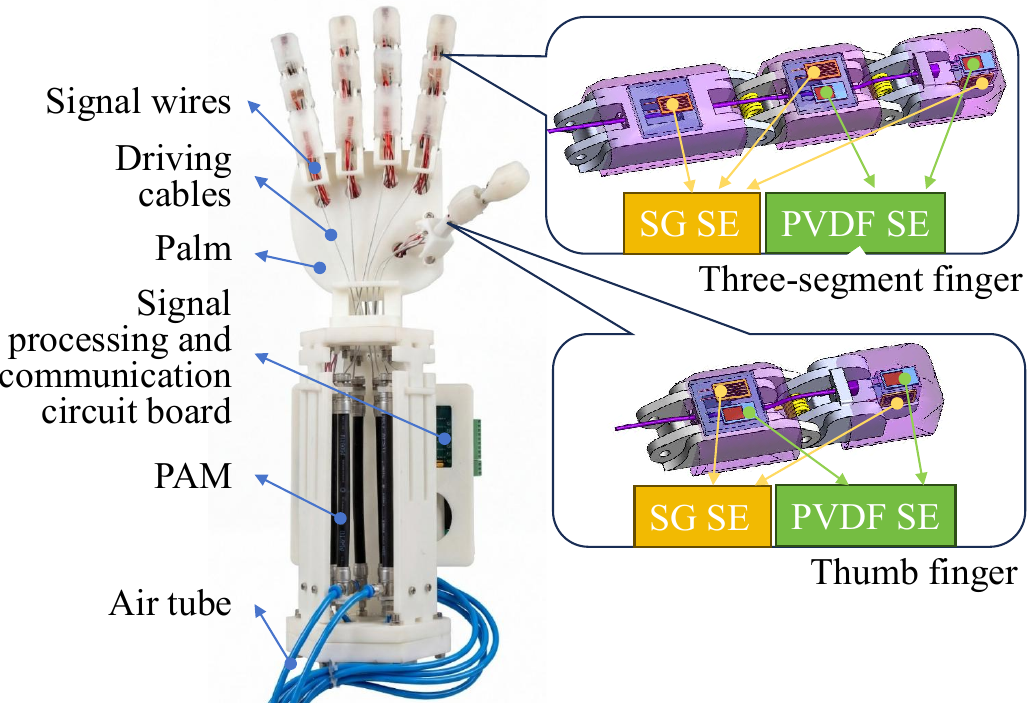}
	\caption{The developed tactile hand and its tactile perception module.}
	\label{fig:structure}
\end{figure}

Bio-inspired by human hand, we designed a five-fingered tactile hand comprising five fingers, a palm and a forearm, as shown in Fig. \ref{fig:structure}. The thumb finger consisted of two segments and the other four fingers three segments. All the fingers were cable-driven and connected to corresponding pneumatic artificial muscle (PAM). To ensure both enough grasp force and flexibility, each finger adopted a soft-rigid-hybrid structure, i.e., rigid phalanx in the center and soft skin in its surrounding, which resembled human finger structure. All tactile SEs were embedded into the soft skin layer.

To simulate the SA and FA mechanoreceptors within human skin, we embedded both piezoresistive and piezoelectric SEs in each finger, which was responsible for the perception of static and dynamic forces respectively \cite{qin2023perception}. Specifically, strain gauge (SG) and PVDF (Polyvinylidene fluoride) were utilized to fabricate the SEs and the detailed manufacturing process could be found in our previous work \cite{shi2023surface}. Different from most existing robotic hands that were equipped with tactile perception only in the fingertip, all the fingers, in our design, were empowered with the capability of multimodal tactile perception by containing both SEs in each finger, as shown in Fig. \ref{fig:structure}. In the two-joint thumb finger, a pair of SG and PVDF was embedded into one finger segment while two pairs of SG and PVDF located in the distal and middle dactylus. Considering the proximal dactylus more possibly touch the palm structure, only one static SE was arranged.

\subsection{Experimental Design}
\subsubsection{Experimental Setup} To collect tactile signals for slip exploration, we set up a robotic manipulation platform as shown in Fig.~\ref{fig:protocol}(a). The tactile hand was installed onto the tip of a robotic arm (Universal Robot Inc., UR5e) which was controlled via ROS (Robot Operating System). The abduction and adduction motion of each finger was driven by an air pump through the PAMs and cables, whose action was controlled by a pneumatic control module and the instruction was sent from a laptop.The tactile signals of each SE channel passed through a preprocessing circuit, where SG-channel signals needed additional conversion from resistor to voltage, and then a multi-channel data acquisition board at 2~KHz sampling rate. The acquired digital signal was transmitted to the laptop. To collect the signals across different materials, six types of materials in total were employed, as shown in Fig. ~\ref{fig:protocol}(b), comprising ABS (Acrylonitrile Butadiene Styrene) plastic, aluminum foil, fiberglass aluminum foil, oil paper, acrylic, and PVC (Polyvinyl chloride). They are indexed as M1 to M6.

\subsubsection{Experimental Process} 
\begin{figure}[htbp]
\centering
\includegraphics[width=\columnwidth]{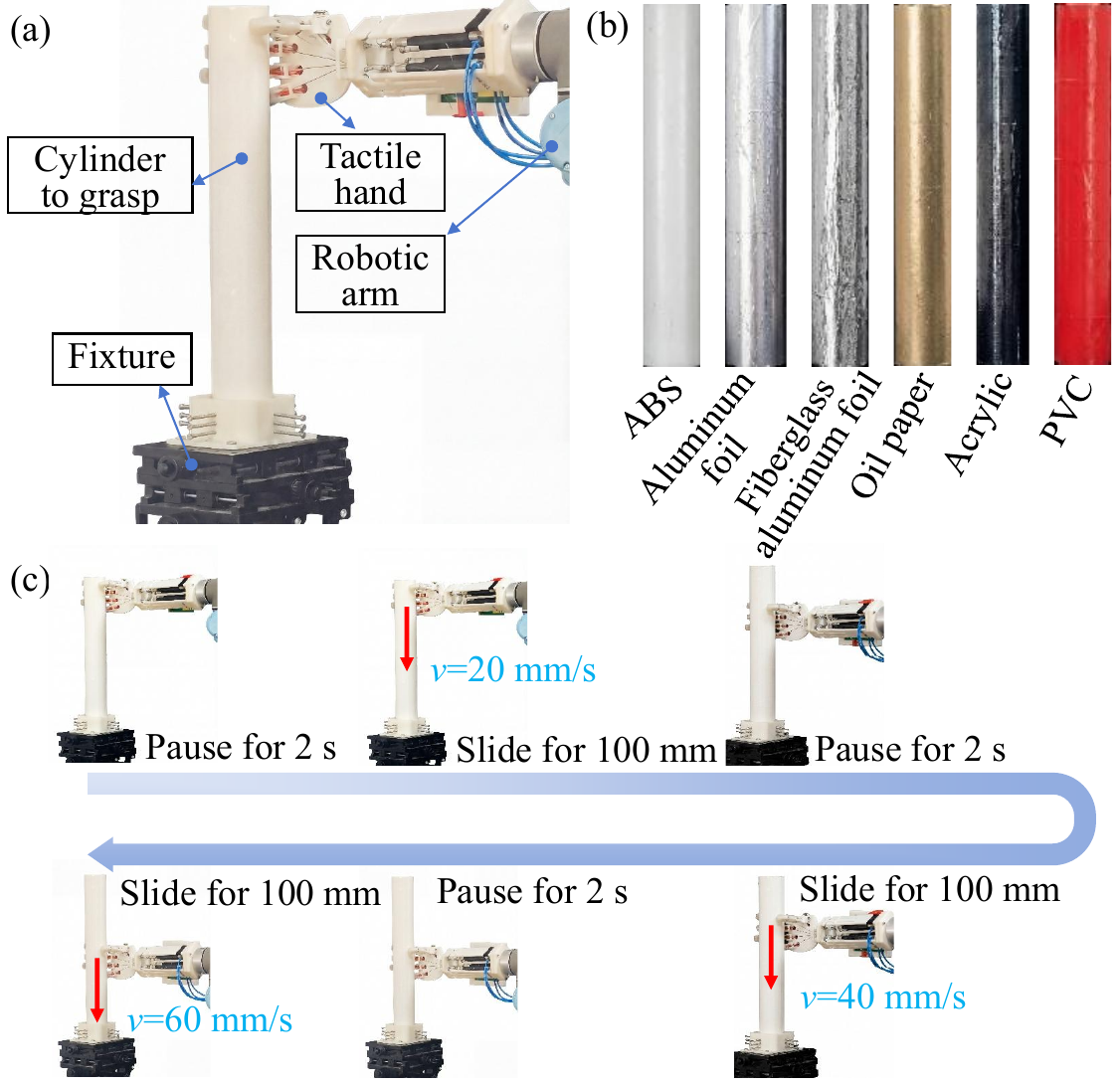}
\caption{Experimental design for the design and signal processing method. (a) Robotic platform. (b) Six materials to explore. (c) Experimental process.}
\label{fig:protocol}
\end{figure}

To generate the tactile signals respectively in static grasping (non-slip) and sliding (slip) period, the cylindrical object was fixed onto a fixture and the tactile hand was controlled to slide vertically along top-down approach. Three sliding velocities were considered. Each experimental cycle was illustrated in Fig. ~\ref{fig:protocol}(c). After 2~s pause, the tactile hand moved downward for 100~mm at 20~mm/s first, and then the velocity changed to 40~mm/s after additional 2~s pause. Finally, the velocity was increased to 60~mm/s. To save time, all these three sub-processes were implemented subsequently and traversed the entire cylinder. The sliding time at each velocity was 5~s, 2.5~s and 1.67~s respectively. As for every material, the experimental cycle was repeated for 28 times. Thus, the experimental dataset was made of six materials and three velocities, each case of which contained 28 trials, producing a total of 504 trials.

\begin{figure*}[hbtp!]
	\centering
	\includegraphics[width=0.9\textwidth]{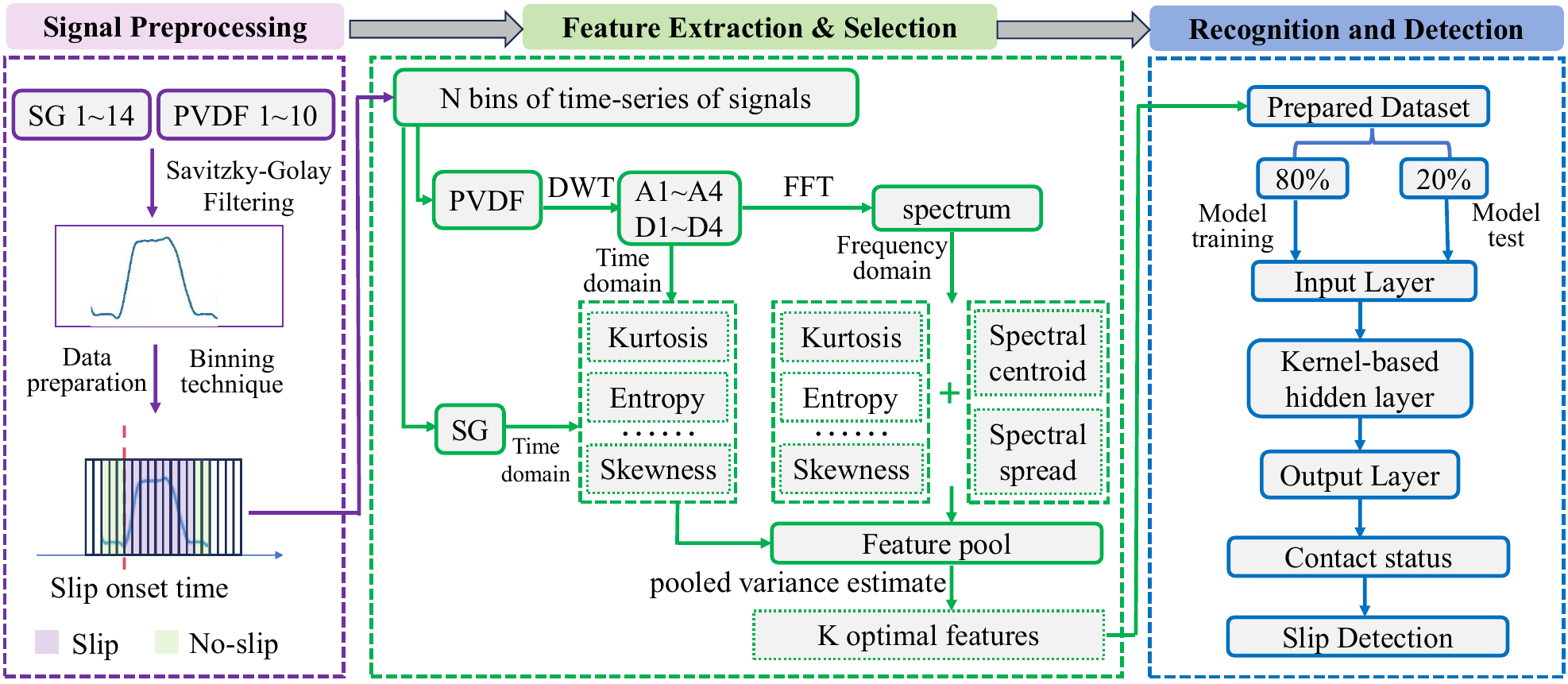}
	\caption{Overall framework of the tactile signal processing approach}
	\label{fig:method}
\end{figure*}

\subsection{Dataset Preparation}
Before feeding to the recognition model, the dataset needed to be segmented and labeled. It was finished manually according to the experimental process. The true slip onset time in each trial was also extracted for each trial. Considering the sliding motion underwent an acceleration, uniform speed and deceleration period, the tactile signals that are 0.5~s ahead of the onset time were labeled as non-slip period while the sliding period that was labeled as slip status started from the true slip onset time and lasted for 0.5~s. 

\section{Methodology}
\subsection{Signal Processing Framework}

The whole signal processing framework was described in Fig.~\ref{fig:method}, which showed it mainly consisted of three procedures: signal preprocessing, feature extraction \& selection, and recognition and detection.

\subsection{Signal Preprocessing}
It was observed that manifest noises existed through the whole experimental process, which might include the interference from surrounding environment, vibrations of the robotic arm and electronic noises from the circuit board etc. Therefore, filtering was necessary as the first step. Here, we applied Savitzky-Golay filter to smooth the tactile signals with the frame length set as 11 and 51 for PVDF and SG respectively. The order was set as 1 for both SEs.

Then, the binning technique was utilized to split the signal period into multiple tiny signal parts. On the one hand, it is aimed at increasing the localization precision for slip onset time. In perfect situations, the first transition from non-slip status to slip can be thought as the exact onset time. On the other hand, it is effective to enlarge the total size of training dataset either for non-slip period or for slip period. Here, we set the bin width to 50~ms after trial-and-error.

\subsection{Feature Extraction and Selection}
In this work, we chose the 17 features following our previous work \cite{zhao2025multi_perspective_feature}, which had shown excellent performance in processing the static and dynamic signals for shape recognition. As for each bin of tactile signals, as shown in Fig.~\ref{fig:method}, both time and frequency domain components were extracted for PVDF-channel signals. Discrete wavelet transform (DWT) was employed to decompose the original signal into four levels of different scales, which were then transformed into frequency domain via Fast Fourier Transform (FFT). Not all channels would be extracted the 17 features. PVDF channels are suitable for 14 and 16 features in time and frequency domain while SG channels are suitable for 13 features in time domain. 

There were a total of 2582 features in the feature pool for each bin of tactile signals. Such high dimensionality is prone to cause the overfitting of a NN and also severely slow down the training speed as a lot of redundancy information is contained. Although neighborhood component analysis had shown outstanding performance in Ref. \cite{zhao2025multi_perspective_feature}, it was not applicable to our case as the capacity of whole feature pool was too large and required significantly increased time cost. Therefore, we employed pooled variance estimate to calculate the importance of each feature. Specifically, the t-test value was computed with pooled variance estimate between every two samples. Once the importance of all features was obtained, we ranked them following the descending order and selected the first K features. $K=120$ was adopted in this work.

\subsection{Recognition and Detection}
Among a variety of NN models, kernel-based extreme learning machine (kernel-ELM) was applied to construct the contact status recognition model due to its rapid computation speed and excellent classification capability. With the conventional back-propagation method and the parameter bias abandoned, ELM calculates the weight matrix quickly via the Moore–Penrose generalized inverse \cite{wang2022review}.

In this work, the polynomial kernel $\kappa(\mathbf{z}_i,\mathbf{z}_j)=(\mathbf{z}_i^\top \mathbf{z}_j + c)^d$ was employed, and $c=0.5$, $d=2$. 80\% of the dataset was used for model training and the rest for testing. The trained model could quickly predict the contact status as non-slip or slip category, according to which, the slip onset time could be determined.

\section{Results}
\subsection{Tactile Signals from Robotic Exploration}

\begin{figure}[htbp]
	\centering
	\includegraphics[width=1\columnwidth]{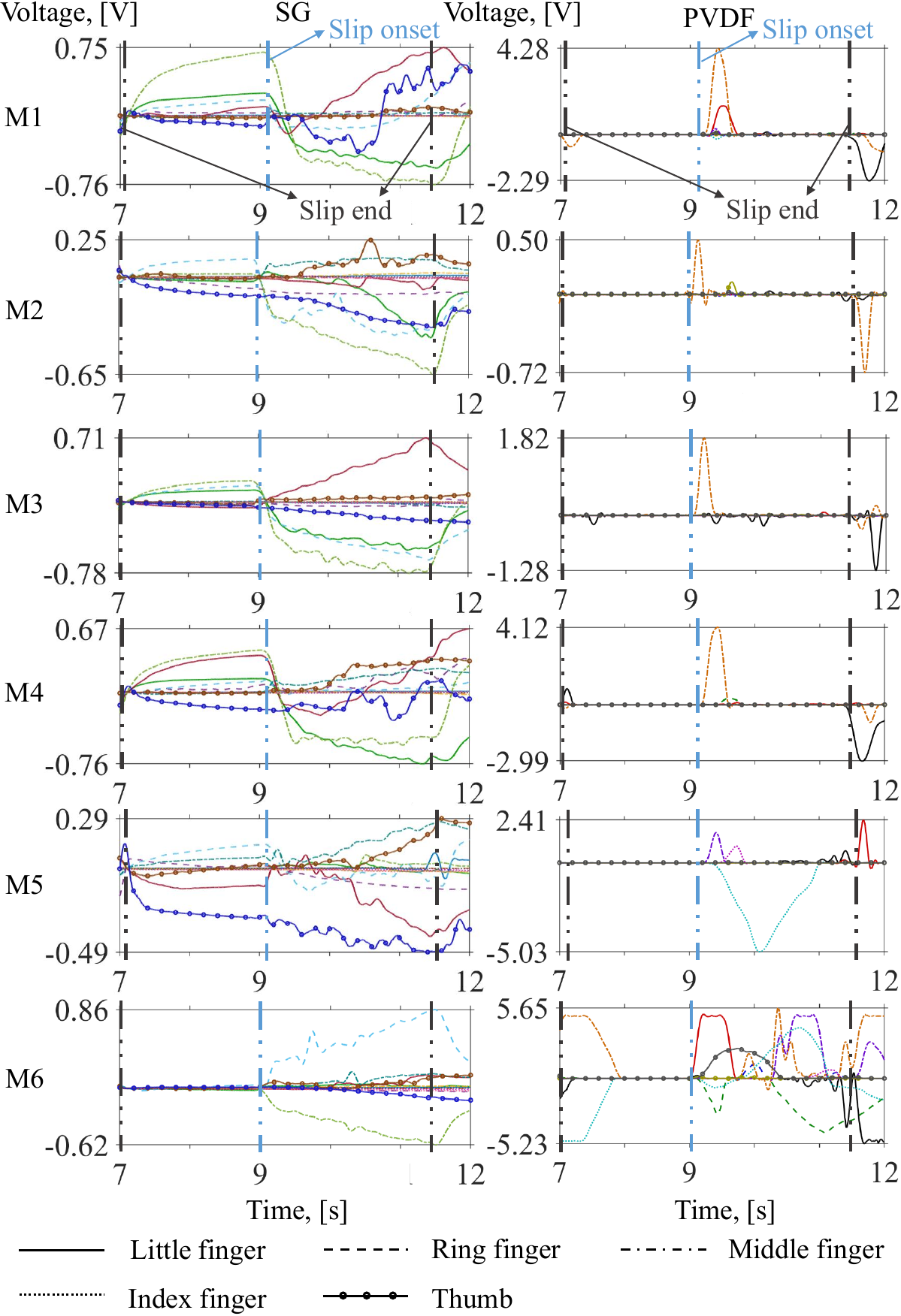}
	\caption{The multi-channel tactile signals generated from the interaction of the tactile hand and grasped objects with the sliding velocity set as 40~mm/s. Different line color represents different SE channel. The left column displays the tactile signals from all 14 SG channels while the right shows the 10 PVDF channels. Each row corresponds to a type of materials (M1$\sim$M6) and different finger is indicated by different line types. The vertical dot dash line in every subplot shows the ground truth of slip onset time.}
	\label{fig:signal_display}
\end{figure}

The acquired tactile signals were shown in Fig.~\ref{fig:signal_display}, where a part of signals near the true slip onset time was displayed lasting for 5 seconds. From the SG-channel signals, it was shown that different magnitude of voltages existed even before the slippage occurred, which might result from two reasons. Firstly, the hand had grasped the object before its slippage and therefore, the voltage during non-slip period didn't stay as zero. Secondly, a variation could be observed since it was a transition period between slip end time and slip onset time. They also indicated that SG channels responded relatively slow and mainly perceive static stimuli. By contrast, no significant variation could be found on PVDF channels during the transition period. A comparison across different materials indicated that this phenomenon behaved distinctively between different materials.

Although the signal magnitude didn't stay the same across different materials, they showed rather similar variation trend in general. While most tactile channels responded to the slippage, a small portion of channels showed inconspicuous response, which accorded with our expectation since only a part of the five fingers contact with the object. After slip onset time, the hand stayed in the sliding period. It was found from the sliding period that most SG channels generated significant variations for all materials with a relative-magnitude-increasing trend. Meanwhile, the PVDF channels mostly underwent abrupt changes instantly after the slip onset time and then recovered to the normal level rapidly for M2$\sim$M5. However, continuous signal variation could be observed for M1 (ABS) and M6 (PVC). It was inferred due to the roughness of these materials, and the roughness of M1 and M6 was higher than the other four materials.

\subsection{Contact Status Recognition}

\begin{figure}[htbp]
	\centering
	\includegraphics[width=0.85\columnwidth]{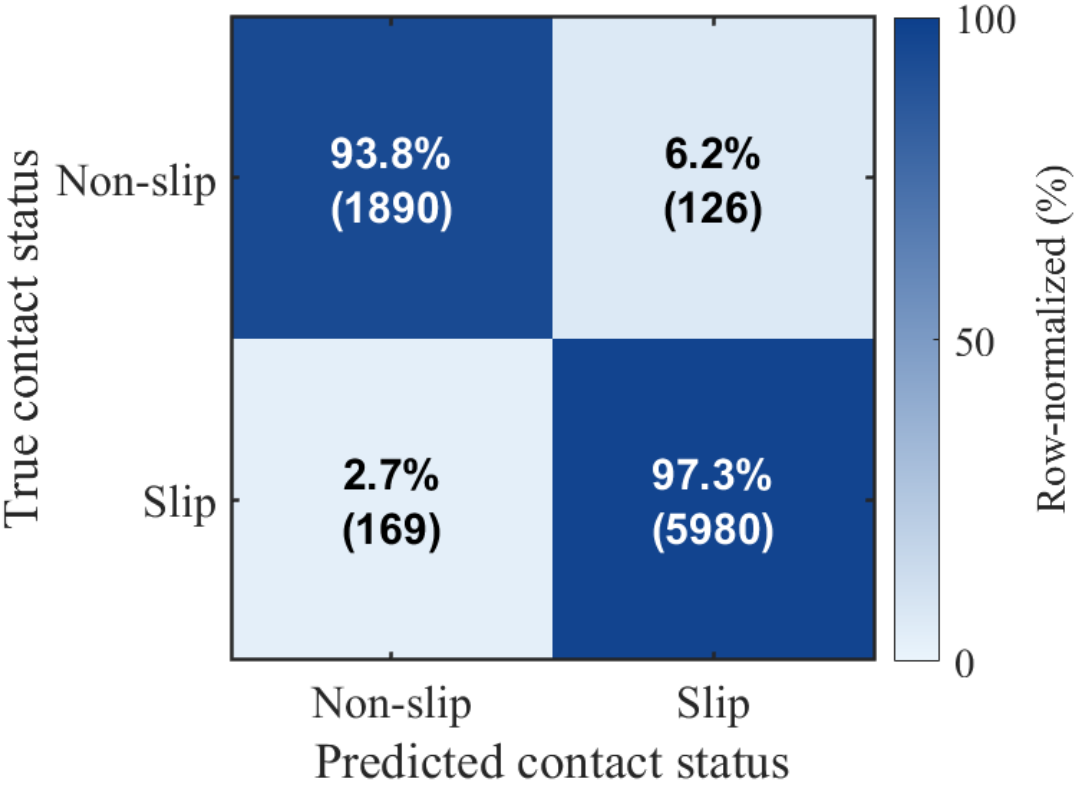}
	\caption{Confusion matrix for the contact status recognition on test dataset of six materials across three sliding velocities.}
	\label{fig:Confusion_matrix_ContactStatus6M}
\end{figure}

When the selected optimal features were fed into the polynomial-kernel ELM model, it finally achieved a high recognition accuracy of 96.59\% and 96.39\% for training and test dataset. The confusion matrix for test data was given in Fig.~\ref{fig:Confusion_matrix_ContactStatus6M}. It could be found 93.8\% and 97.3\% of binned tactile signals were correctly recognized as non-slip and slip status respectively. At the same time, non-slip status was more likely to be misrecognized than slip in general.

\begin{figure}[htbp]
	\centering
	\includegraphics[width=1\columnwidth]{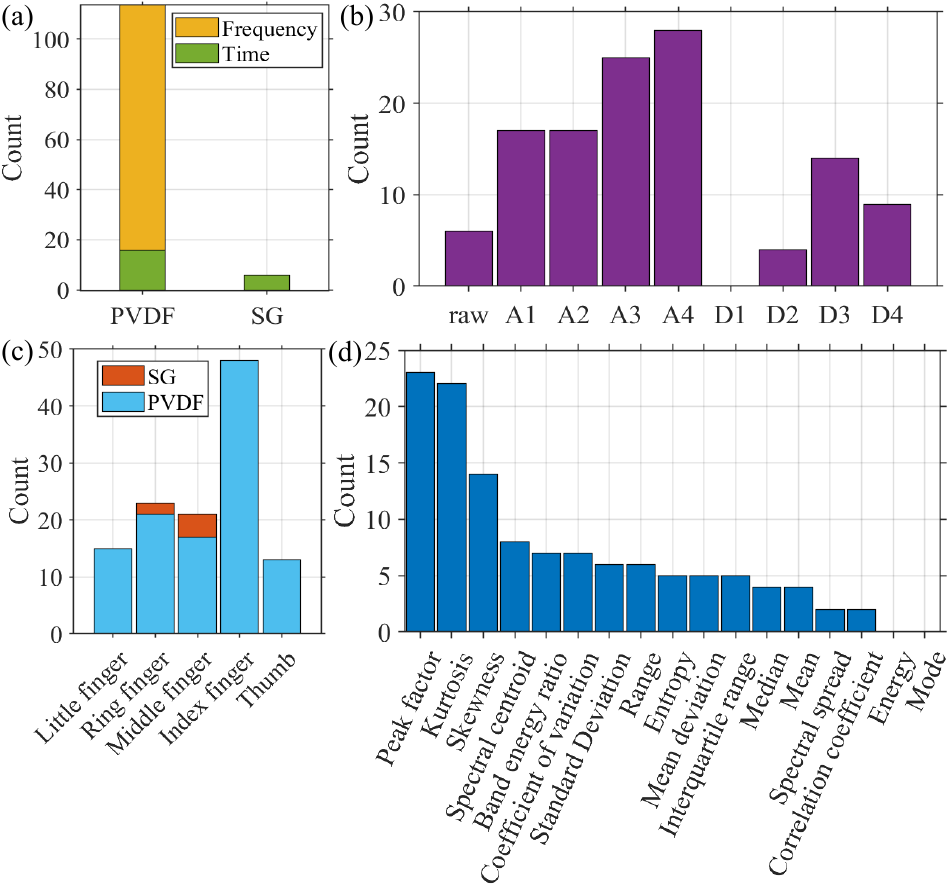}
	\caption{Statistical characteristics of the 120 optimal features selected by the  pooled variance estimate algorithm. (a)$\sim$(d) respectively show the feature count versus SE type, DWT component, Finger type, and feature definitions.}
	\label{fig:Feature_statistics}
\end{figure}

Meanwhile, the characteristics of 120 selected features were summarized in Fig.~\ref{fig:Feature_statistics} in order to reveal the significance of different factors in the recognition of contact status. It revealed from Fig.~\ref{fig:Feature_statistics}(a) that PVDF SE played a more evident role than SG, and frequency domain generated more selected features than time domain. Among the different signal components, approximation levels (A1$\sim$A4) seemed more crucial than detail levels (D1$\sim$D4) and no feature from D1 components was employed. It is interesting in Fig.~\ref{fig:Feature_statistics}(c) that not all SGs on the five fingers were employed. A small amount of SG channels was chosen only on the ring and middle finger.The majority of features selected by the pooled variance estimate were generated on the index finger.Fewer features were utilized on the little and thumb finger. Among the 17 types of feature definitions, only 15 features were employed and the top three features were peak factor, kurtosis and skewness, whose counts were much larger than the others.

\subsection{Generalization to Unseen Materials}
Although it has shown excellent performance in contact status recognition across six materials and three sliding velocities. It still remains unknown whether the trained recognition model stays valid in contact status recognition when objects of novel unseen materials are manipulated. The property of material is closely related with the viscosity, roughness and friction force of the object surface, and thus, it directly influences the slippage phenomenon between robotic hand and manipulated object. The applicability would be severely restricted if re-training is required for new materials. Therefore, it is essential that the trained recognition model remains effective for unseen novel materials.

\begin{figure}[htbp]
	\centering
	\includegraphics[width=1\columnwidth]{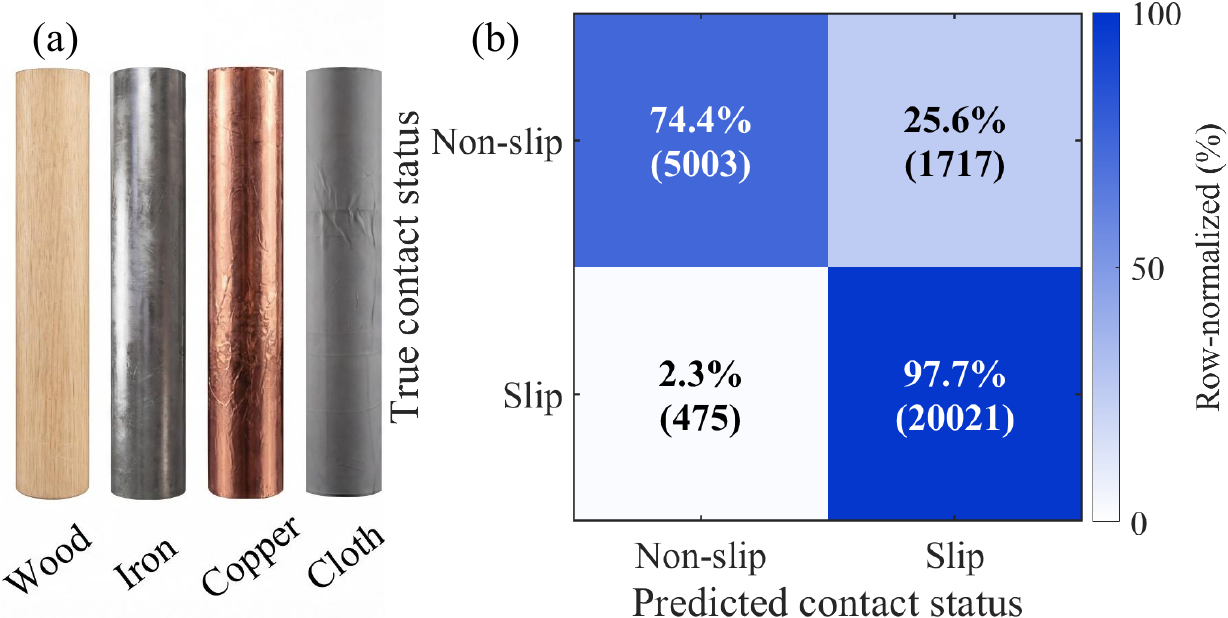}
	\caption{Four unseen materials to verify the generalization ability of the contact status recognition (a) and the confusion matrix of the recognition result (b). The overall recognition accuracy was 91.95\%.}
	\label{fig:Confusion_matrix_CSUnseen4M}
\end{figure}

In this work, we verified our trained model in the preceding subsection on four novel unseen materials as shown in Fig.~\ref{fig:Confusion_matrix_CSUnseen4M}(a), without retraining or any fine-tuning. The four materials were Wood, Iron, Copper, Cloth. The recognition result was given in Fig.~\ref{fig:Confusion_matrix_CSUnseen4M}(b). The overall recognition accuracy was decreased to 91.95\%, whose performance was degraded compared with the recognition on trained materials. But it still maintained within a relatively high level. As shown in Fig.~\ref{fig:Confusion_matrix_CSUnseen4M}(b), it could be found that 74.4\% and 97.7\% of non-slip and slip status were identified correctly. It revealed that the recognition of non-slip status was more challenging since some involved noises or transition phases interfered the recognition.

\subsection{Slip Detection based on the Recognition Result}
\begin{figure}[htbp]
	\centering
	\includegraphics[width=1\columnwidth]{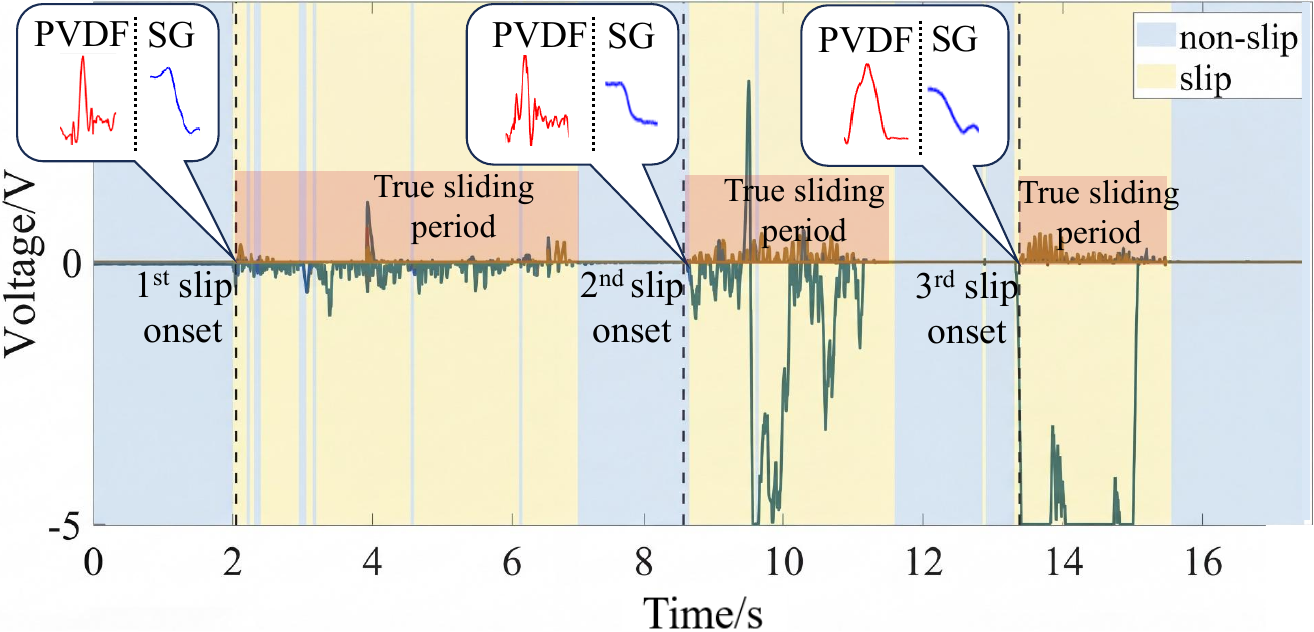}
	\caption{Recognition result of the contact status on the material M3, Fiberglass aluminum foil. It comes from one original trial of PVDF-channel signals which was generated from three sliding velocities: 20~mm/s, 40~mm/s, and 60~mm/s. The blue and yellow color indicated non-slip and slip status.}
	\label{fig:Slip_Detection}
\end{figure}

Further, we tested the slip detection effect of our method based on the contact status recognition result. The original tactile signal was utilized as test data, which comprised different sliding velocities and no signal partition was applied. One trial of the PVDF-channel tactile signals on Fiberglass aluminum foil was shown in Fig.~\ref{fig:Slip_Detection}, considering PVDF channels played a more significant role in the status recognition.

Due to the difference of three sliding velocities, the signal period decreases over time and the amplitude also differed. Higher sliding speed tended to produce larger amplitude. In general, the tactile signal during slip period exhibited a completely different variation trend, showing apparent fluctuations. The recognized status was also shown in the form of bins in Fig.~\ref{fig:Slip_Detection} with blue and yellow color indicating the non-slip and slip contact status. It was observed that most contact status was recognized correctly with a small portion of binned signals was misrecognized. The slip onset time could be detected according to the transition from non-slip to slip status. However, some incorrect detection events would also occur since the recognition accuracy of contact status was less than 100\%. To increase the detection accuracy, more endeavors could be devoted to the improvement of recognition accuracy.

\section{Conclusion}
In this work, we presented how to realize the contact status recognition and slip detection with our developed five-fingered tactile hand. From the perspective of feature engineering, all 24 channels of tactile signals were proposed with binning technique applied. In the case of our trained material, an accuracy as high as 96.39\% was achieved. The selected 120 optimal features were also analyzed further in view of the SE type, signal component, finger type and feature definitions. When applied to four unseen material, the accuracy also reached 91.95\%. Finally, we found it had shown valuable potential in slip detection. Currently, perfect detection cannot be realized based on mere recognition of contact status, which limitation could be addressed in future by improving the recognition accuracy and introducing a possibility model for slip detection.

\bibliographystyle{IEEEtran}
\bibliography{references}

@article{roberts2021soft,
  title={Soft tactile sensing skins for robotics},
  author={Roberts, Peter and Zadan, Mason and Majidi, Carmel},
  journal={Current Robotics Reports},
  volume={2},
  number={3},
  pages={343--354},
  year={2021},
  publisher={Springer}
}

@article{zhan2025recent,
  title={Recent advances and challenges of tactile sensing for robotics: from fundamentals to applications},
  author={Zhan, Ziheng and Yang, Yang and Zuo, Wenjuan and Xie, Mingzhu and Ning, Meng},
  journal={Materials Today Physics},
  volume={54},
  pages={101740},
  year={2025},
  publisher={Elsevier}
}

@article{xin2025vision_based_tactile,
  title={Vision-based tactile sensing: From performance parameters to device design},
  author={Xin, Yi Hang and Hu, Kai Ming and Xiang, Rui Jia and Gao, Yu Ling and Zhou, Jun Feng and Meng, Guang and Zhang, Wen-Ming},
  journal={Applied Physics Reviews},
  volume={12},
  number={2},
  year={2025},
  publisher={AIP Publishing}
}

@article{li2020skin_inspired_quadruple,
  title={Skin-inspired quadruple tactile sensors integrated on a robot hand enable object recognition},
  author={Li, Guozhen and Liu, Shiqiang and Wang, Liangqi and Zhu, Rong},
  journal={Science Robotics},
  volume={5},
  number={49},
  pages={eabc8134},
  year={2020},
  publisher={American Association for the Advancement of Science}
}

@article{xu2024selfpowered_eskin,
  title={Self-powered flexible electronic skin tactile sensor with 3D force detection},
  author={Liu, Jize and Zhao, Wei and Ma, Zhichao and Zhao, Hongwei and Ren, Luquan},
  journal={Materials Today},
  volume={81},
  pages={84--94},
  year={2024},
  publisher={Elsevier}
}

@article{lambeta2020digit,
  title={Digit: A novel design for a low-cost compact high-resolution tactile sensor with application to in-hand manipulation},
  author={Lambeta, Mike and Chou, Po-Wei and Tian, Stephen and Yang, Brian and Maloon, Benjamin and Most, Victoria Rose and Stroud, Dave and Santos, Raymond and Byagowi, Ahmad and Kammerer, Gregg and others},
  journal={IEEE Robotics and Automation Letters},
  volume={5},
  number={3},
  pages={3838--3845},
  year={2020},
  publisher={IEEE}
}

@article{sui2022incipient,
  title={Incipient slip detection method for soft objects with vision-based tactile sensor},
  author={Sui, Ruomin and Zhang, Lunwei and Li, Tiemin and Jiang, Yao},
  journal={Measurement},
  volume={203},
  pages={111906},
  year={2022},
  publisher={Elsevier}
}

@inproceedings{wang2021gelsight,
  title={Gelsight wedge: Measuring high-resolution 3d contact geometry with a compact robot finger},
  author={Wang, Shaoxiong and She, Yu and Romero, Branden and Adelson, Edward},
  booktitle={2021 IEEE international conference on robotics and automation (ICRA)},
  pages={6468--6475},
  year={2021},
  organization={IEEE}
}

@article{zhao2025multi_perspective_feature,
  title={Tactile Exploration Enabled Shape Recognition with Multi-Perspective Feature Representation},
  author={Zhao, Hongliang and Shi, Xiaowei and Yang, Wenhui and Chen, Huayang and Qin, Longhui},
  journal={IEEE Sensors Journal},
  year={2025},
  publisher={IEEE}
}

@inproceedings{egli2024sensorized,
  title={Sensorized soft skin for dexterous robotic hands},
  author={Egli, Jana and Forrai, Benedek and Buchner, Thomas and Su, Jiangtao and Chen, Xiaodong and Katzschmann, Robert K},
  booktitle={2024 IEEE International Conference on Robotics and Automation (ICRA)},
  pages={18127--18133},
  year={2024},
  organization={IEEE}
}

@article{min2025stretchable,
  title={A stretchable tactile sensor with deep learning-enabled 3D force decoding for human and robotic interfaces},
  author={Min, Shunhua and Geng, Haoyang and He, Yuheng and Liang, Wensheng and Chen, Shoubin and Wang, Zhijun and Liu, Qingzhou and Xu, Tailin},
  journal={Chemical Engineering Journal},
  pages={167189},
  year={2025},
  publisher={Elsevier}
}

@article{park2025manufacturing,
  title={Manufacturing strategies for highly sensitive and self-powered piezoelectric and triboelectric tactile sensors},
  author={Park, Hyosik and Gbadam, Gerald Selasie and Niu, Simiao and Ryu, Hanjun and Lee, Ju-Hyuck},
  journal={International Journal of Extreme Manufacturing},
  volume={7},
  number={1},
  pages={012006},
  year={2025},
  publisher={IOP Publishing}
}

@article{agriomallos2018slippage,
  title={Slippage detection generalizing to grasping of unknown objects using machine learning with novel features},
  author={Agriomallos, Ioannis and Doltsinis, Stefanos and Mitsioni, Ioanna and Doulgeri, Zoe},
  journal={IEEE Robotics and Automation Letters},
  volume={3},
  number={2},
  pages={942--948},
  year={2018},
  publisher={IEEE}
}

@article{chen2025learning,
  title={Learning-based Slip Detection and Fine Control Using the Tactile Sensor for Robot Stable Grasping},
  author={Chen, Zhangyi and Wang, Long and Luo, Yao and Li, Xiaoling and Li, Shuai},
  journal={IEEE Robotics and Automation Letters},
  year={2025},
  publisher={IEEE}
}

@article{funabashi2022gcn_tactile,
  title={Multi-fingered in-hand manipulation with various object properties using graph convolutional networks and distributed tactile sensors},
  author={Funabashi, Satoshi and Isobe, Tomoki and Hongyi, Fei and Hiramoto, Atsumu and Schmitz, Alexander and Sugano, Shigeki and Ogata, Tetsuya},
  journal={IEEE Robotics and Automation Letters},
  volume={7},
  number={2},
  pages={2102--2109},
  year={2022},
  publisher={IEEE}
}

@article{chen2025crossmodal_slip,
  title={Temporal cross-modal fusion method of tactile, 2D visual, and point cloud data for slip detection in robotic grasping},
  author={Chen, Kaichao and Chen, Gang and Lin, Jirong and Gao, Xianyuan and Zhao, Xiangrong},
  journal={Measurement Science and Technology},
  volume={36},
  number={10},
  pages={106220},
  year={2025},
  publisher={IOP Publishing}
}

@inproceedings{jawale2024learned,
  title={Learned slip-detection-severity framework using tactile deformation field feedback for robotic manipulation},
  author={Jawale, Neel and Kaur, Navneet and Santoso, Amy and Hu, Xiaohai and Chen, Xu},
  booktitle={2024 IEEE/RSJ International Conference on Intelligent Robots and Systems (IROS)},
  pages={13569--13576},
  year={2024},
  organization={IEEE}
}

@article{hu2024learning,
  title={Learning to detect slip through tactile estimation of the contact force field and its entropy properties},
  author={Hu, Xiaohai and Venkatesh, Aparajit and Wan, Yusen and Zheng, Guiliang and Jawale, Neel and Kaur, Navneet and Chen, Xu and Birkmeyer, Paul},
  journal={Mechatronics},
  volume={104},
  pages={103258},
  year={2024},
  publisher={Elsevier}
}

@inproceedings{gao2024aim_bionic_tactile,
  title={A solid-liquid composite flexible bionic tactile sensor for dexterous hands},
  author={Gao, Zheng and Gong, Zhenhua and Zhu, Guangpu and Zhang, Ting},
  booktitle={2024 IEEE International Conference on Advanced Intelligent Mechatronics (AIM)},
  pages={1560--1566},
  year={2024},
  organization={IEEE}
}

@article{qin2023perception,
  title={Perception of static and dynamic forces with a bio-inspired tactile fingertip},
  author={Qin, Longhui and Shi, Xiaowei and Wang, Yihua and Zhou, Zhitong},
  journal={Journal of Bionic Engineering},
  volume={20},
  number={4},
  pages={1544--1554},
  year={2023},
  publisher={Springer}
}

@article{shi2023surface,
  title={Surface recognition with a bioinspired tactile fingertip},
  author={Shi, Xiaowei and Wang, Yihua and Qin, Longhui},
  journal={IEEE Sensors Journal},
  volume={23},
  number={16},
  pages={18842--18855},
  year={2023},
  publisher={IEEE}
}

@article{wang2022review,
  title={A review on extreme learning machine},
  author={Wang, Jian and Lu, Siyuan and Wang, Shui-Hua and Zhang, Yu-Dong},
  journal={Multimedia Tools and Applications},
  volume={81},
  number={29},
  pages={41611--41660},
  year={2022},
  publisher={Springer}
}

@article{romeo2020slip_survey,
  title={Methods and sensors for slip detection in robotics: A survey},
  author={Romeo, Rocco A and Zollo, Loredana},
  journal={Ieee Access},
  volume={8},
  pages={73027--73050},
  year={2020},
  publisher={IEEE}
}

@article{qin2025recent,
  title={Recent progress and challenges of key technologies in robotic assembly},
  author={Qin, Longhui},
  journal={Chinese Journal of Mechanical Engineering},
  pages={100032},
  year={2025},
  publisher={Elsevier}
}

@article{romeo2021automatic_slippage,
  title={Method for automatic slippage detection with tactile sensors embedded in prosthetic hands},
  author={Romeo, Rocco A and Lauretti, Clemente and Gentile, Cosimo and Guglielmelli, Eugenio and Zollo, Loredana},
  journal={IEEE Transactions on Medical Robotics and Bionics},
  volume={3},
  number={2},
  pages={485--497},
  year={2021},
  publisher={IEEE}
}

@article{liu2024ultrasensitive,
  title={Ultrasensitive touch sensor for simultaneous tactile and slip sensing},
  author={Liu, Yue and Tao, Juan and Mo, Yepei and Bao, Rongrong and Pan, Caofeng},
  journal={Advanced Materials},
  volume={36},
  number={21},
  pages={2313857},
  year={2024},
  publisher={Wiley Online Library}
}

\end{document}